# Implementation of an Innovative Bio Inspired GA and PSO Algorithm for Controller design considering Steam GT Dynamics

R.Shivakumar  and  Dr. R.Lakshmipathi

[1] Department of Electrical and Electronics Engineering
Sona College of Technology, Salem-636005, INDIA

[2] Department of Electrical and Electronics Engineering
St.Peters Engineering College, Chennai, INDIA.

**Abstract**
The Application of Bio Inspired Algorithms to complicated Power System Stability Problems has recently attracted the researchers in the field of Artificial Intelligence. Low frequency oscillations after a disturbance in a Power system, if not sufficiently damped, can drive the system unstable. This paper provides a systematic procedure to damp the low frequency oscillations based on Bio Inspired Genetic (GA) and Particle Swarm Optimization (PSO) algorithms. The proposed controller design is based on formulating a System Damping ratio enhancement based Optimization criterion to compute the optimal controller parameters for better stability. The Novel and contrasting feature of this work is the mathematical modeling and simulation of the Synchronous generator model including the Steam Governor Turbine (GT) dynamics. To show the robustness of the proposed controller, Non linear Time domain simulations have been carried out under various system operating conditions. Also, a detailed Comparative study has been done to show the superiority of the Bio inspired algorithm based controllers over the Conventional Lead lag controller.

## 1. Introduction

Modern Bio Inspired Algorithms include a wide variety of Population based algorithms which can be applied to various Power System Optimization problems. The Phenomenon of stability of modern interconnected power systems has received a great deal of attention in recent years. One problem that faces power systems nowadays is the Low frequency oscillations arising from interconnected Power Systems [1]. Sometimes, these oscillations sustain for minutes and grow to cause system separation, if adequate damping is not provided. A cost efficient and satisfactory solution to the problem of low frequency oscillations is to provide damping by implementing Power System Stabilizers (PSS), which are supplementary controllers in the Generator Excitation Systems [2-3].Designing and applying the PSS is a complex phenomenon. In recent years, several approaches based on Modern control theory have been applied to PSS design problem. These include optimal control, adaptive control, variable structure control and intelligent control [4-5].

Despite the potential of modern control techniques, Power System utilities still prefer the conventional lead lag PSS structure [6].The parameters of CPSS are based on a linearized model of the Power System. Since modern power systems are dynamic and non linear in nature, CPSS performance is degraded whenever the operating point changes from one point to another because of the fixed parameters of the stabilizer. Unfortunately, the conventional techniques are time consuming, as they are iterative and require heavy computation burden and slow convergence.

Recently, Bio Inspired optimization techniques like Genetic Algorithm, Evolutionary Programming, Simulated Annealing, Bacteria foraging, Particle Swarm optimization etc have been applied for PSS parameter optimization [7-9].In this work, Genetic algorithm and Particle Swarm Optimization algorithms has been implemented for computing the parameters of the optimal controller including Steam Governor Turbine dynamics for Power System Stability.

 The Main Objectives of this work are summarized as follows:
(1).To develop a linearized State Space model of the SMIB model (including Steam Governor Turbine dynamics) with and without the damping controller.

(2).To formulate a Damping ratio enhancement based optimization criterion to compute the Optimal PSS parameters.

(3). To compare the computed damping ratios of the Conventional PSS (CPSS), Genetic based PSS (GAPSS) and Particle Swarm based PSS(PSOPSS) for damping the poorly damped Electromechanical modes of oscillation.

(4). To carry out Parameter Sensitivity analysis by performing Non linear Time domain based simulation to validate the robustness of the proposed controllers under wide variations of system operating conditions and also under variations in system parameter involved in the model.

## 2. Modeling of Power System including GT Dynamics

2.1 System Model under Study

Fig.1.shows the Heffron Phillips block diagram [10] of Single Machine Infinite bus model (SMIB) equipped with PSS. EXC(s) in the Heffron model represents the IEEE Type 1 Excitation system involving Amplifier, Exciter and Rate Derivative feedback compensation as in Fig(2).

In all the classical model analysis for SMIB system, the mechanical power input remains constant during the period of the Transient (i.e) Governor Turbine Dynamics is not included in the modeling and analysis. But in this work, the mechanical power input is included in the Modeling and Simulation. PSS in the model represents the Power System Stabilizer with $\Delta\omega$ as the input and output is $\Delta U_E$ given to the Generator Excitation System summing point.

Fig3.represents the Steam Governor Turbine Model with Time Constants $T_{RH}$, TCH, and $T_{GV}$.
Where  $T_{RH}$ = Reheater Time Delay.
  $T_{CH}$ = Inlet and Steam Chest Delay.
  $T_{GV}$ = Main gate Servo Motor Time Constant
  $R_P$ = Steady State Speed Droop.
  $F_{HP}$ = High Pressure (HP) flow fraction

In this work, Non Reheat type Steam Turbine is used in the modeling and simulation.

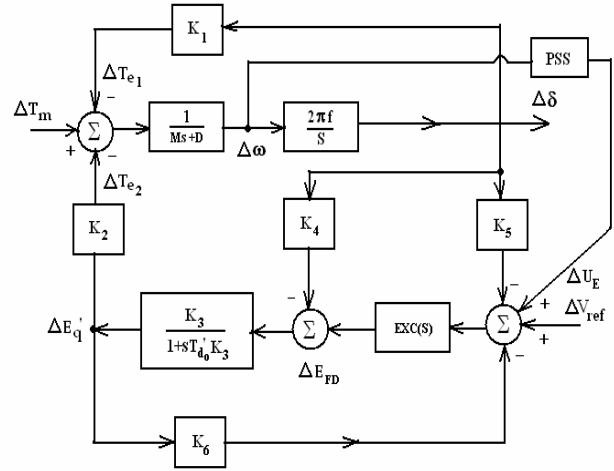

Fig.1. Heffrons-Phillips SMIB Model with PSS

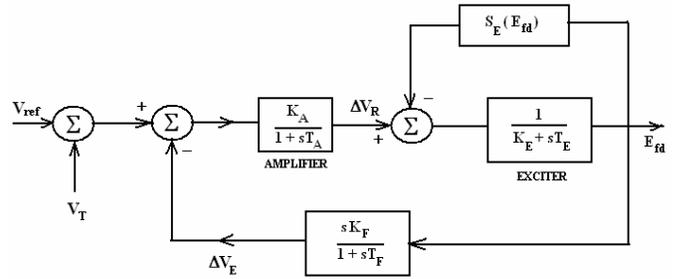

Fig.2. IEEE type 1 Excitation System Model

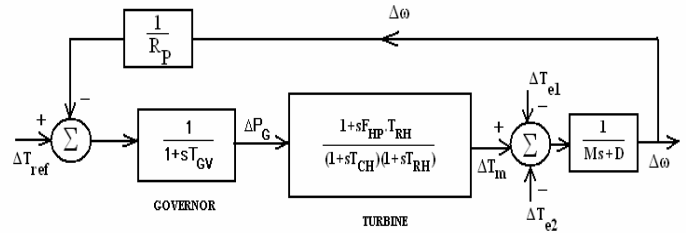

Fig.3. Steam Governor Turbine Model

For a Non Reheat Steam Turbine, $T_{RH}$ =0, where
$T_{RH}$ = Reheater Delay (Typically 6 secs).
Hence the model in Fig (3) is simplified to the Model as in Fig (4).

In Fig (4), $T_{GS}$ = Steam Governor Time Constant
$T_{TS}$ = Steam Turbine Time Constant.

Fig(4) represents the model for Steam Governor and Turbine(Non reheat type) to be equipped with the Heffron Phillips Generator model.(i.e) Output($\Delta Tm$) of Steam GT Model is given as input to the Heffron-Phillips generator model.
All the abbreviations for the Constants and Variables involved in the model are given in Appendix-II.

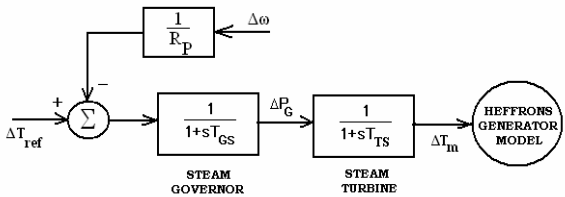

Fig.4. Steam Governor-Non Reheat type Turbine Model

The Dynamic Model in state space is given by

$$\dot{x} = Ax + Bu \qquad (1)$$

where [x] = Vector of State Variables
A,B = State Matrix and Input Matrix respectively.
In this work, for Open loop 8 state variables and for Closed loop (with PSS) 11 state variables are used in the modeling.

$$[x]_{open} = [\Delta\omega, \Delta\delta, \Delta Eq', \Delta E_{FD}, \Delta V_R, \Delta V_E, \Delta P_G, \Delta Tm]^T \qquad (2)$$

$$[x]_{Closed} = [\Delta\omega, \Delta\delta, \Delta Eq', \Delta E_{FD}, \Delta V_R, \Delta V_E, \Delta P_G, \Delta Tm, \Delta P_1, \Delta P_2, \Delta U_E]^T \qquad (3)$$

All the Test System parameters used for simulation [11] are given in Appendix-I.

## 2.2. Structure of PSS

The PSS model consists of the Gain Block, Cascaded identical Phase Compensation block and the washout block. The input to the controller is the Rotor speed Deviation ($\Delta\omega$) and output is the Supplementary Control signal ($\Delta U_E$) given to Generator excitation system.

The Transfer function of the PSS Model is given by

$$\left[\frac{\Delta U_E}{\Delta\omega}\right] = Ks\left[\frac{(1+sT_1)}{(1+sT_2)}\right]\left[\frac{(1+sT_3)}{(1+sT_4)}\right]\left[\frac{(sT_w)}{(1+sT_w)}\right] \qquad (4)$$

Where Ks = PSS gain
Tw = Washout Time Constant
$T_1, T_2, T_3, T_4$ = PSS Time Constants
Time Constants $T_1=T_3$, $T_2=T_4$
(Identical Compensator Block).

Hence Ks, $T_1$, $T_2$ are the PSS parameters which should be computed using Conventional Lead Lag stabilizer and optimally tuned using GAPSS and PSOPSS. The washout time constant (Tw) is in the range of 1 to 20 seconds and in this work, Tw is taken as 10 seconds.

## 3. Proposed Optimization Criterion for Damping

### 3.1 Criteria for Damping

The Rate of Decay of amplitude of oscillations is best expressed in terms of the Damping ratio ($\xi$). For an Oscillatory mode represented by an Complex Eigen value ($\sigma \pm j\omega$), the Damping ratio is given by

$$[\xi] = \frac{-\sigma}{\sqrt{\sigma^2 + \omega^2}} \qquad (5)$$

The damping ratio of all system modes of oscillation should exceed a specified value. In Power Systems, Electromechanical oscillations with damping ratios more than 0.05 are considered satisfactory.

### 3.2 Proposed Optimization Criterion

The Main objective of this formulation is to compute the optimal value of PSS parameters for system oscillations damping. A good measure of system damping is the damping ratio or damping factor. Hence a Damping Ratio enhancement based objective function is selected for PSS parameter optimization.

$$[J] = \min(\xi_i), \xi_i \varepsilon \xi_{EM} \qquad (6)$$

where $\xi_i$ = Damping Ratio of $i^{th}$ Electromechanical Mode of Oscillation.

$\xi_{EM}$ = Damping Ratios of all the Electromechanical modes of Oscillation.

The Objective here is to Maximize J, in order to enhance the Damping Ratio of the poorly damped modes of oscillation for better Stability.

The poorly damped electromechanical modes of oscillation will have its Eigen values located in right half of complex s plane, thus making the system Unstable.

The above criterion formulation is represented in simpler form as:

Maximize [J] such that

$$\left[\min(\xi_i)\right] \xrightarrow{Shift} \left[\xi_{sm}\right], \xi_{sm} \geq \xi_T \quad (7)$$

Where  $\xi_T$ = Threshold Level of Damping ratio for System Stability.
  $\min(\xi_i)$ = Minimum Damping ratio value among the Electromechanical Modes of Oscillation.
  $\xi_{sm}$ = Required Damping ratio for Stability with m = 1, 2, 3.

Here m represents the method used for optimization. (i.e) m=1 indicate CPSS, m=2 indicate GAPSS, m=3 indicate PSOPSS.

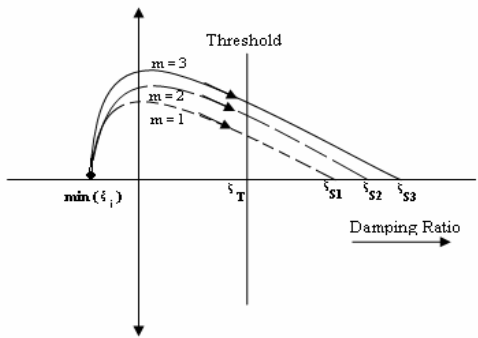

Fig.5. Pictorial Representation of the proposed Objective function.

In Fig.5, Damping ratio corresponding to min($\xi_i$) location is to be shifted to locations beyond $\xi_T$ for system stability. Let it be $\xi_{S1}, \xi_{S2}, \xi_{S3}$ for the three techniques implemented in this work (CPSS, GAPSS, PSOPSS).

The Design problem including the constraints imposed on the various PSS parameters is given as follows:

**Optimize J**
Subject to
$$K_s^{\min} \leq K_s \leq K_s^{\max} \quad (8)$$
$$T_1^{\min} \leq T_1 \leq T_1^{\max} \quad (9)$$
$$T_2^{\min} \leq T_2 \leq T_2^{\max} \quad (10)$$

Typical ranges selected for Ks, $T_1$ and $T_2$ are as follows: For Ks [5 to 60], for $T_1$ [0.1 to 1] and for $T_2$ [0.1 to 1].

This Maximization criterion has been implemented in this work to compute the Optimum Value of the PSS Parameters Ks, $T_1$ and $T_2$.

## 4. Bio Inspired Algorithms

4.1 Genetic Algorithm- An Overview.

Genetic Algorithms are numerical optimization algorithms inspired by Natural selection and Natural Genetics [12-13]. GA techniques differ from more traditional search algorithms in that they work with a number of candidate solutions rather than one candidate solution. Each candidate solution of a problem is represented by a data structure known as an individual. A group of individuals collectively comprise what is known as a population. GAs are initialized with a population of random guesses. GA includes operators such as Reproduction, Crossover, Mutation and Inversion.

*Reproduction* is a process in which a new generation of population is formed by selecting the fittest individuals in the current population. *Crossover* is responsible for producing new offsprings by selecting two strings and exchanging portions of their structures. The new offsprings may replace the weaker individuals in the population.

*Mutation* is a local operator which is applied with a very low probability of occurrence. Its function is to alter the value of a random position in a string.

Finally, *Inversion* is a process which inverts the order of the elements between two randomly chosen points on the string.

The Algorithmic Steps involved in Genetic Algorithm are as follows:

Step 1. Specify the various parameters for GA Optimization.

Step 2. Create an Initial Population of individuals randomly.

Step 3. Evaluate the Fitness of each individual (i.e) Evaluating the optimization criterion J.

Step 4. If value of J obtained is minimum, then Optimum value of PSS parameters is equal to those obtained in current generation, Otherwise Goto step 5.

**Step 5.** Based on the fitness, select the best Individuals and perform recombination through a crossover process.

**Step 6.** Mutate the new generation with a given Probability.

**Step 7.** If termination condition (Maximum no of Generations) is not reached, go back to step (3).

4.2 Particle Swarm Optimization- An Overview.

PSO is an Evolutionary Computation Technique developed by Eberhart and Kennedy [14-15] in 1995, which was inspired by the Social behavior of Bird flocking and fish schooling. PSO has its roots in artificial life and social psychology as well as in Engineering and Computer science [16-17].It is not largely affected by the size and Non linearity of the problem and can converge to the optimal solution in many problems where most analytical methods fail to converge.

Particle Swarm Optimization has more advantages over Genetic Algorithm as follows:

(a). PSO is easier to implement and there are fewer parameters to adjust.
(b). In PSO, every particle remembers its own previous best value as well as the neighbourhood best ; therefore, it has a more effective memory capability than GA.
(c). PSO is more efficient in maintaining the diversity of the swarm, since all the particles use the information related to the most successful particle in order to improve themselves, whereas in Genetic algorithm, the worse solutions are discarded and only the new ones are saved; (i.e) in GA the population evolves around a subset of the best individuals.

PSO utilizes a population of particles that fly through the problem space with given velocities. Each particle has a memory and it is capable of remembering the best position in the search space ever visited by it. The Positions corresponding to the Best fitness is called *Pbest* and the overall best out of all the particles in the population is called *gbest.*
At each iteration, the velocities of the individual particles are updated according to the best position for the particle itself and the neighborhood best position.

The velocity of each agent can be modified by the following equation

$$V_i^{k+1} = W \cdot V_i^k + C_1 \cdot rand_1 * (Pbest_i - S_i^k) + C_2 \cdot rand_2 * (g_{best} - S_i^k) \quad (11)$$

Where
- $V_i^k$ = Velocity of agent i at iteration k.
- $W$ = Weighting Function.
- $C_j$ = Weighting factor.
- $rand$ = random number between 0 and 1.
- $S_i^k$ = Current position of agent i at iteration k.
- $Pbest$ = Pbest of agent i.
- $gbest$ = gbest of the group.

The following Weighting Function is usually utilized in equation (11).

$$W = [W_{max}] - \left[\frac{W_{max} - W_{min}}{iter_{max}}\right] * iter \quad (12)$$

where
- $W_{max}$ = Initial Weight
- $W_{min}$ = Final Weight.
- $iter_{max}$ = Maximum Iteration number
- $iter$ = Current iteration number.

The Current position can be modified by the following equation

$$S_i^{k+1} = S_i^k + V_i^{k+1} \quad (13)$$

The Algorithmic Steps involved in Particle Swarm Optimization Algorithm are as follows:

**Step 1:** Select the various parameters of PSO.

**Step 2:** Initialize a Population of particles with random Positions and Velocities in the problem space.

**Step 3** : Evaluate the Desired Optimization Fitness Function for each particle.

**Step 4:** For each Individual particle, Compare the Particles fitness value with its Pbest. If the Current value is better than the pbest value, then set this value as the Pbest for agent i.

**Step 5:** Identify the particle that has the Best Fitness Value. The value of its fitness function is identified as gbest.

**Step 6:** Compute the new Velocities and Positions of the particles according to equation (11) & (13).

**Step 7**: Repeat steps 3-6 until the stopping Criterion of Maximum Generations is met.

## 5. Simulation Results

For all the Computation, Simulation and Analysis of the results in this work, MATLAB 7.0 / SIMULINK platform was used.
The Two main analysis involved in the simulation in this work are
(1). Small Signal Stability Analysis.
(2). Non Linear Time Domain Analysis.

(1).Small Signal Stability Analysis.

Small Signal Stability is the ability of the Power System to maintain synchronism when subjected to small disturbances.In this work, the stability analysis is based on computation of eigen values and damping ratios of the system for open loop, with CPSS, GAPSS and PSOPSS and its comparison.
The disturbances involved are variation in system operating point with Load change disturbance, variation in system parameters namely variation in line reactance, variation in Amplifier Gain with respect to the normal operating point.

The Eigen values located in the left half of complex s-plane will determine the stability of the system, whereas the Eigen values located in right half of complex s-plane will make the system unstable.The Damping ratios more than the threshold value of damping will provide better system damping.

(2).Non linear Time Domain Analysis.

This analysis is to show the effectiveness of the proposed controllers in damping the low frequency oscillations under wide variations in operating conditions and system parameters.

The objective is to minimize the integral squared error (ISE) involved in the system. The error refers to the speed deviation ($\Delta\omega$) and the power angle deviation ($\Delta\delta$).

The Integral squared error is given by,

$$[ISE] = \int_0^{Ts} e^2(t)dt \qquad (14)$$

Here e(t) refers to the error involving Rotor Speed deviation ($\Delta\omega$) and Power Angle deviation ($\Delta\delta$).Ts represent the Time of Simulation.

The State Space modeling of the SMIB model including steam Governor Turbine dynamics was performed and the system open loop eigen values and damping ratios was computed as listed in Table 1 and Table 2.The Electromechanical modes of oscillation indicate that the test system is unstable having positive real part eigen values located in right half of the complex s plane.

Also, the time domain analysis involving Load change disturbance ($\Delta P_L$) in Fig (6) and Fig (7) reveal that the open loop system without PSS are having oscillatory responses with huge overshoots and large settling times, thus making the system unstable.

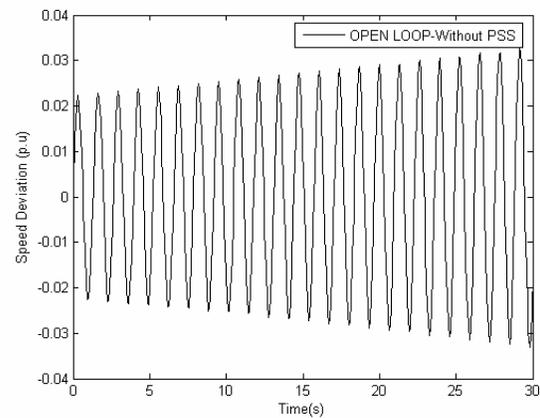

Fig.6.Open Loop Speed Deviation Response at (0.4, 0.008), $\Delta P_L$ =0.1p.u Operating Condition.

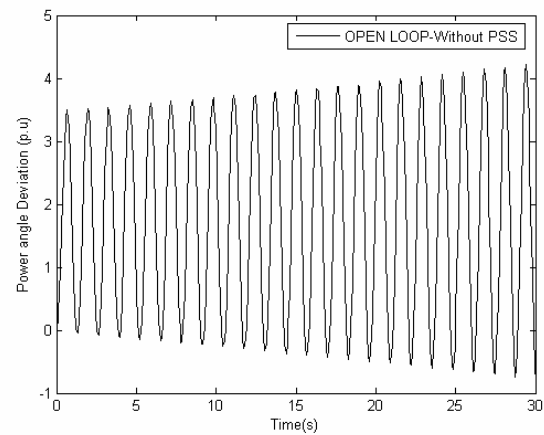

Fig.7.Open Loop Power Angle Deviation Response at (0.4, 0.008), $\Delta P_L$=0.1p.u Operating Condition

Implementation of CPSS, GAPSS and PSOPSS provide the Optimal PSS parameters as listed in Table 2.

Table 1. Computed Eigen Values for Open Loop without PSS, GAPSS and PSOPSS

| S. No | Operating Conditions | Eigen Values | | | |
|---|---|---|---|---|---|
| | | Open Loop without PSS | CPSS | GAPSS | PSOPSS |
| 1 | P = 0.4<br>Q = 0.008<br>$\Delta P_L$ = 0.1 p.u | -13.1440<br>**0.1218 ± j 5.452**<br>-6.5964<br>-3.2215 ± j 4.7092<br>-2.0101<br>-1.3302 | -16.7162 ± j 7.1748<br>**-0.3956 ± j 8.6327**<br>-0.8315 ± j 3.4119<br>-4.950 ± j 0.7994<br>-0.0500<br>-2.2377<br>-1.3260 | -15.3069 ± j 5.6824<br>**-0.7011 ± j 7.2045**<br>-1.3476 ± j 3.9158<br>-5.4757 ± j 1.0014<br>-0.0501<br>-2.2702<br>-1.3303 | -15.8761 ± j 6.8962<br>**-0.8895 ± j 8.4064**<br>-5.8273 ± j 0.7754<br>-2.1101 ± j 2.4145<br>-0.0504<br>-1.6176<br>-1.8079 |
| 2 | P = 0.4<br>Q = 0.06<br>$\Delta P_L$ = 0.2 p.u | -13.1452<br>**0.1231 ± j 5.405**<br>-6.6024<br>-3.2247 ± j 4.7140<br>-2.0000<br>-1.3292 | -16.6293 ± j 7.0840<br>**-0.4355 ± j 8.5255**<br>-0.8501 ± j 3.4180<br>-4.9452 ± j 0.8198<br>-0.0501<br>-2.2343<br>-1.3233 | -14.8144 ± j 5.1911<br>**-0.5502 ± j 6.7845**<br>-1.6839 ± j 3.9568<br>-5.7962 ± j 0.9729<br>-0.0501<br>-2.2542<br>-1.3364 | -15.7799 ± j 4.7389<br>**-0.6361 ± j 7.0777**<br>-2.9594 ± j 2.6991<br>-0.0504<br>-1.6220 ± j 0.2209<br>-2.2438<br>-1.3253 |
| 3 | P = 0.4<br>Q = 0.06<br>$\Delta P_L$ = 0.3 p.u<br>+ 10 % increase in Line reactance Xe. | -13.0586<br>**0.1272 ± j 5.6292**<br>-6.5683<br>-3.2617 ± j 4.6158<br>-2.0552<br>-1.3371 | -17.1748 ± j 7.6603<br>**-0.2107 ± j 9.1457**<br>-0.7241 ± j 3.3516<br>-4.7499 ± j 0.6580<br>-0.0500<br>-2.2364<br>-1.3329 | **-0.3440 ± j 5.8868**<br>-12.1442 ± j 1.6734<br>-3.2217 ± j 4.0608<br>-9.0214<br>-6.2420<br>-0.0501<br>-1.3637<br>-2.2169 | -16.7640 ± j 7.1869<br>**-0.5372 ± j 8.7165**<br>-0.7460 ± j 3.5708<br>-4.8116 ± j 0.7338<br>-0.0500<br>-2.2373<br>-1.3334 |

Table 2. Computed Damping Ratios for Open loop without PSS, GAPSS and PSOPSS

| S. No | Operating Conditions | Optimal Damping Controller Parameters | | | Damping Ratios of Poorly Damped Electromechanical Modes of Oscillation<br>Threshold Level of Damping Ratio ($\xi_T$)= 0.06 | | | |
|---|---|---|---|---|---|---|---|---|
| | | CPSS $[K_s,T_1,T_2]$ | GAPSS $[K_s,T_1,T_2]$ | PSOPSS $[K_s,T_1,T_2]$ | Open Loop Without PSS | CPSS | GAPSS | PSOPSS |
| 1 | P = 0.4<br>Q = 0.008<br>$\Delta P_L$ = 0.1 p.u | 6.1692<br>0.6707<br>0.1000 | 6.2634<br>0.4557<br>0.5823 | 52.1596<br>0.2353<br>0.5176 | -0.02233 | 0.04578 | 0.09693 | **0.105225** |
| 2 | P = 0.4<br>Q = 0.06<br>$\Delta P_L$ = 0.2 p.u | 6.2986<br>0.6487<br>0.1000 | 8.1981<br>0.3527<br>0.8915 | 43.2273<br>0.1605<br>0.3724 | -0.022769 | 0.051016 | 0.08083 | **0.089513** |
| 3 | P = 0.4,<br>Q = 0.06,<br>$\Delta P_L$ = 0.3 p.u<br>+ 10 % increase in Line reactance Xe. | 5.1944<br>0.8100<br>0.1000 | 8.7912<br>0.1359<br>0.2346 | 14.9908<br>0.4651<br>0.1521 | -0.022591 | 0.023032 | 0.05834 | **0.061513** |

Table.3. Parameters selected for GA Implementation.

| GA Parameters | |
|---|---|
| Population Size | 20 |
| No of Generations | 10 |
| Selection Operator | Roulette Wheel Selection |
| Generation gap | 0.9 |
| Crossover Probability | 0.95 |
| Mutation Probability | 0.10 |
| Termination Method | Maximum Generations |

Table.4. Parameters selected for PSO Implementation.

| PSO Parameters | |
|---|---|
| Swarm Size | 20 |
| $w_{max}, w_{min}$ | 1 , 0.5 |
| $C_1, C_2$ | 1.0 , 1.0 |
| No of Generations | 10 |
| No of Variables | 03 |
| Termination Method | Maximum Generations |

Table 3, Table 4 represents the various parameters selected for Bio Inspired algorithms implementation.
**Parameter Sensitivity Analysis:**

Parameter Sensitivity analysis refers to analyzing the System Stability performance, whenever the System parameters involved in the System model and System operating conditions are varied over a wide range with respect to the normal operating point.

In this work, the system is subjected to wide variations as follows.

(a). Wide variation in operating condition(P,Q) with Load change disturbances ($\Delta P_L$) introduced in the system model.
(b). 10 % variation (increase) in Line Reactance with respect to the normal operating point.
(c). 10 % variation (increase) in Amplifier gain $K_A$ with respect to the normal operating point.

In Power Systems, Electromechanical oscillations with damping ratios greater than 0.05 are considered satisfactory [18]. Based on this criterion, the desired threshold level of damping is taken as $\xi_T = 0.06$ in this work.

The Damping Ratios are computed for the poorly damped Electromechanical modes of oscillation using the optimal PSS parameters and eigen values, listed in Table 2. The Damping ratios calculated for poorly damped modes of oscillation reveal that the proposed controllers provide better damping to the oscillatory modes. Though CPSS and GAPSS provide good damping, the PSO based controller (PSOPSS) provide better damping to the oscillatory modes, with damping ratios more than the threshold level of damping ($\xi_T=0.06$) for all the conditions involved (last Column of Table 2).

Non linear Time domain simulations involving wide variations in operating points and system model parameters have been carried out to show the robustness of the proposed controllers in damping the low frequency oscillations.

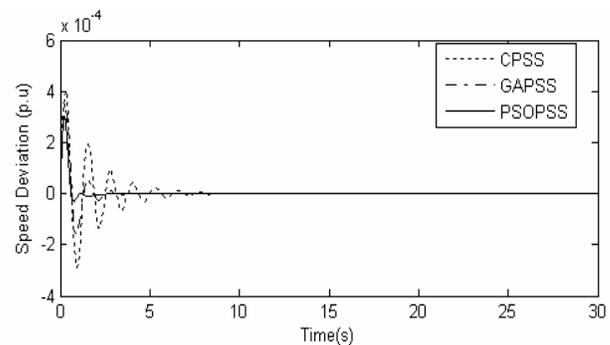

Fig.8. Speed Deviation response for (0.4, 0.008), $\Delta P_L$=0.1p.u condition with CPSS, GAPSS and PSOPSS.

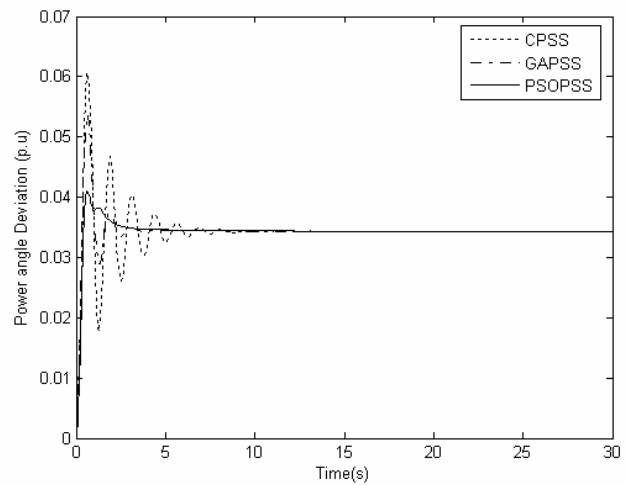

Fig.9. Power Angle Deviation response for (0.4, 0.008), $\Delta P_L$=0.1p.u condition with CPSS, GAPSS and PSOPSS

Fig (8) and Fig (9) shows the effectiveness of the Bio inspired controllers in damping the low frequency oscillations better than the CPSS for the operating condition (P=0.4, Q=0.008, Load Change disturbance of 0.1 p.u).

Fig (10) and Fig (11) indicate the Speed deviation and Power Angle response of the system with operating condition (P=0.4, Q= 0.06, $\Delta P_L$=0.2 p.u).

These responses reveal the dominance of the Bio Inspired optimal damping controllers in damping out the low frequency oscillations, in particular, the PSO based controller damp the oscillations with reduced overshoot and quick settling time compared to the CPSS and the Genetic based PSS (GAPSS).

compared to the CPSS and the Genetic based PSS (GAPSS).

In order to enhance the system stability, the CPSS, GAPSS and PSOPSS reduces the oscillations overshoot and also make the oscillations to settle at a quicker settling time. For instance, in Fig (13), the maximum overshoot is 0.06 p.u for CPSS, for GA based PSS it is 0.054p.u, whereas for PSOPSS, the maximum oscillation overshoot is only 0.038 p.u. This shows the optimal tuning and effective damping exerted by the PSOPSS.

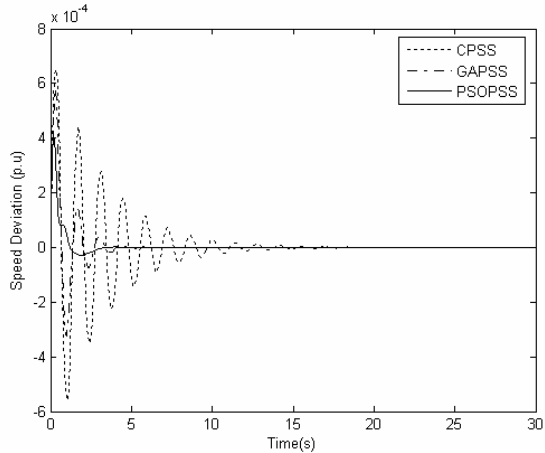

Fig.10. Speed Deviation response for (0.4, 0.06), $\Delta P_L$=0.2 p.u condition with CPSS, GAPSS and PSOPSS

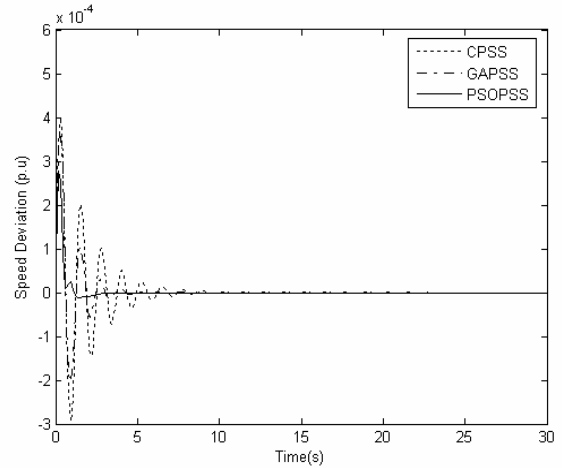

Fig.12. Speed Deviation response for (0.4, 0.06), $\Delta P_L$=0.3 p.u and 10% increase in Line Reactance(Xe) condition with CPSS, GAPSS and PSOPSS.

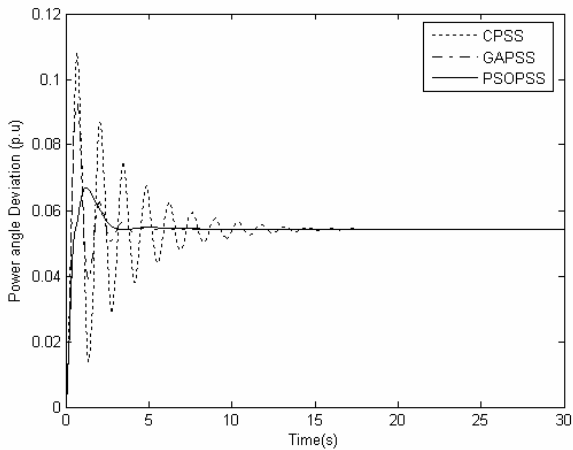

Fig.11. Power Angle Deviation response for (0.4, 0.06),$\Delta P_L$=0.2 p.u Condition with CPSS, GAPSS and PSOPSS.

Fig (10) and Fig (11) indicate the Speed deviation and Power Angle response of the system with operating condition (P=0.4, Q= 0.06, $\Delta P_L$=0.2 p.u).These responses reveal the dominance of the Bio Inspired optimal damping controllers in damping out the low frequency oscillations, in particular, the PSO based controller damp the oscillations with reduced overshoot and quick settling time

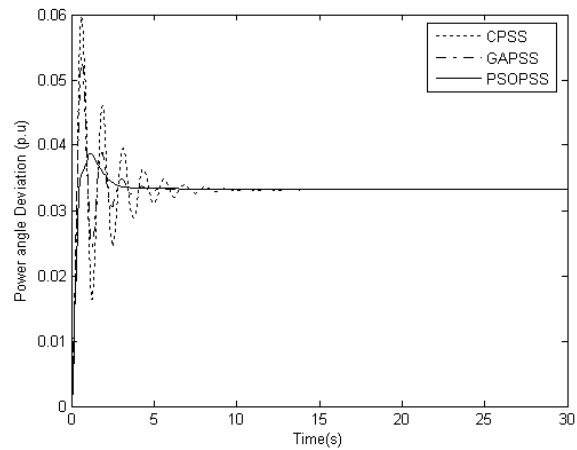

Fig.13. Power Angle Deviation response for (0.4, 0.06), $\Delta P_L$=0.3 p.u and 10% increase in Line Reactance (Xe) condition with CPSS, GAPSS
  and PSOPSS.

These responses clearly indicate the damping phenomenon exerted by the various controllers. In all the cases, the Bio Inspired algorithms(GA and PSO) based controller provide better damping to the Low frequency oscillations compared to the conventional lead lag stabilizer,thus enhancing Power system stability .

## 6. Conclusions

This work provides a better and efficient solution to the complicated Engineering optimization problem of damping the low frequency oscillations, by implementing the GA and PSO algorithms, thus enhancing Power System Stability.

The following are the important features implemented satisfactorily in this work:

**(a).** State Space Modeling of the SMIB System including the Governor Turbine(Non Reheat Type) dynamics for analysis and Simulation.

**(b).** Implementation of Damping Ratio based optimization criterion to compute the optimal PSS parameters based on Conventional, Genetic and PSO based algorithms.

**(c).** Carried out a detailed comparative study on the damping ratios for CPSS, GAPSS and PSOPSS to damp out the poorly damped electromechanical modes of oscillation.

**(d).** Implementation of Parameter Sensitivity analysis to validate the robustness of the proposed controllers by performing Non linear Time domain based simulations under wide system loading conditions and also under various system parameter variations.

### Appendix – I

**Test System Parameters**

Generator : $x_d$ = 0.973, $x_d'$=0.190, $x_q$ = 0.550,
D=0, M= 9.26, $T_{do}'$ = 7.76 secs
Excitation : IEEE ST1A type Excitation
(For Speed input Stabilizer)
$K_A$ = 190, $T_A$ =0, $K_F$ = 0, $T_F$ =1Sec

Line and Load: R = 0.034, Xe = 0.997, G = 0.249,
B= 0.262, $V_{to}$ = 1.05, = 0.4, Q = 0.008.
Governor and
Turbine : Steam Type
$T_{RH}$=0 (Non Reheat Type), $R_T$ = 0.4,
$R_P$ = 0.05, $T_{GS}$ = 0.2, $T_{TS}$ = 0.3,
$T_{CH}$ = 0.3 Secs.
All Parameters are in p.u unless specified otherwise.

### Appendix –II

**Nomenclature**

$\Delta\omega$ = Incremental Change in Rotor Speed
$\Delta\delta$ = Incremental Change in Rotor Power Angle.
$\Delta E_q'$ = Incremental Change in Generator Internal Voltage.
$\Delta E_{FD}$ = Incremental Change in Generator Field Voltage.
$\Delta V_R$ = Incremental Change in Amplified Voltage.
$\Delta V_E$ = Incremental Change in Rate Feedback Compensation output voltage.
$S_E(E_{fd})$ = Excitation Saturation Function.
$K_F$ = Gain of Rate Feedback compensation
$T_F$ = Time Constant of Rate Feedback Compensation
$K_E$ = Gain of Exciter
$T_E$ = Time Constant of Exciter
$K_A$ = Gain of Amplifier
$T_A$ = Time Constant of Amplifier.
$T_{do}'$ = Field Open circuit Time constant
$\Delta P_1$ = Output State Variable of PSS Washout Block.
$\Delta P_2$ = Output State variable of first PSS Phase Compensation block
$\Delta P_G$ = Incremental Change in Generation (Output State Variable of Steam Governor)
$\Delta U_E$ = Supplementary Excitation signal from PSS.
$K_1$-$K_6$ = K Constants/ Coefficients involved in Heffrons Phillips Model.
$\xi$ = System Damping ratio
$\xi_T$ = Threshold level of Damping ratio.
$\sigma$ = Real part of Eigen value
$\omega$ = Imaginary part of Eigen Value
Xe = Transmission Line Reactance
$\Delta P_L$ = Load Change Disturbance
M = Inertia Constant
D = System Damping
Tw = Washout Time Constant
Ks = Gain of Power System Stabilizer
ISE = Integral Squared Error.
CPSS = Conventional Lead Lag Stabilizer.
GAPSS = Genetic Algorithm based PSS.
PSOPSS= Particle Swarm based PSS.
SMIB = Single Machine Infinite Bus System.
$V_T$ = Generator Terminal Voltage
$\Delta T_{ref}$ = Incremental Change in Reference Torque

**R.Shivakumar** received the M.E degree in Power System Engineering from Annamalai University, India in 1999. Currently he is working in Sona College of Technology, Salem, India as Assistant Professor in Electrical and Electronics Engineering Department. He is now working towards his PhD in Computational Intelligence at Anna University, Chennai. His areas of research include Artificial Intelligence, Optimization Techniques, FACTS and Power System Stability Analysis. He is an annual member of IEEE and Life member of ISTE.

**Dr.R.Lakshimpathi** received the B.E degree in 1971 and M.E degree in 1973 from College of Engineering, Guindy, and Chennai.He received his PhD degree in High Voltage Engineering from Indian Institute of Technology, Chennai, India. He has 36 years of teaching experience in various Government Engineering Colleges in Tamilnadu and he retired as Principal and Regional Research Director at Alagappa Chettiar College of Engineering and Technology, Karaikudi.He is now working as Professor in Electrical and Electronics Engineering department at St.Peters Engineering College, Chennai.His areas of research include HVDC Transmission, Power System Stability and Electrical Power Semiconductor Drives.